\pdfoutput=1
\documentclass[a4paper]{llncs}

\usepackage{amssymb}
\usepackage{xcolor}
\usepackage{graphicx}
\usepackage{subfig}
\usepackage{multirow}
\usepackage{array}
\usepackage{authblk}
\usepackage{booktabs}
\newcolumntype{C}[1]{>{\centering\arraybackslash\hspace{0pt}}p{#1}}
\usepackage{url}
\urldef{\mailsa}\path|{alfred.hofmann, ursula.barth, ingrid.haas, frank.holzwarth,|
\urldef{\mailsb}\path|anna.kramer, leonie.kunz, christine.reiss, nicole.sator,|
\urldef{\mailsc}\path|erika.siebert-cole, peter.strasser, lncs}@springer.com|    

\urldef{\mails}\path|{woalsdnd, khwan.jung}@vuno.co, sangjunpark@snu.ac.kr| 

\begin{document}

\mainmatter  

\title{Classification of Findings with Localized Lesions in Fundoscopic Images using a Regionally Guided CNN}

\author{Jaemin Son$^1$ \and Woong Bae$^1$ \and Sangkeun Kim$^1$ \and Sang Jun Park$^2$ \and Kyu-Hwan Jung$^1$\footnote{Corresponding author}}
\institute{$^1$VUNO Inc., Seoul, Korea,\\
\path|{woalsdnd, iorism, sisobus, khwan.jung}@vuno.co|\\
$^2$Department of Ophthalmology, Seoul National University College of Medicine,\\ Seoul National University Bundang Hospital, Seongnam, Korea\\
\path|sangjunpark@snu.ac.kr|}


%
%

\maketitle

\begin{abstract}
Fundoscopic images are often investigated by ophthalmologists to spot abnormal lesions to make diagnoses. Recent successes of convolutional neural networks are confined to diagnoses of few diseases without proper localization of lesion. In this paper, we propose an efficient annotation method for localizing lesions and a CNN architecture that can classify an individual finding and localize the lesions at the same time. Also, we introduce a new loss function to guide the network to learn meaningful patterns with the guidance of the regional annotations. In experiments, we demonstrate that our network performed better than the widely used network and the guidance loss helps achieve higher AUROC up to $4.1\%$ and superior localization capability.
\end{abstract}

\section{Introduction}
Fundoscopic images provide comprehensive visual clues about the condition of the eyes. For the analysis, ophthalmologists search for abnormal visual features called {\it findings} from the images and make decisions on {\it diagnoses} based on the findings discovered. For instance, the severity of diabetic retinopathy (DR) is clinically judged by the existence and the extent of relevant findings (microaneurysm, hemorrhage, hard exudate and cotton wool patch, etc)~\cite{wilkinson2003proposed}.

In recent years, Convolutional Neural Networks (CNN) have achieved the level of professional ophthalmologists in diagnosing DR and diabetic macular edema (DME)~\cite{gulshan2016development,ting2017development}. However, CNNs in the literature are trained to make decisions on the diagnoses directly without localizing lesions. There exist several studies that visualize the lesions that contribute to the decision on diagnoses~\cite{gargeya2017automated,takahashi2017applying}, though, types of findings were not identified for the lesions.

In the past, segmentation methods with hand-crafted feature-extractors had been proposed for the detection of hemorrhage~\cite{bae2011study}, hard Exudate~\cite{sasaki2013quantitative}, drusen deposits~\cite{rapantzikos2003detection} and cotton wool patch~\cite{kose2012simple}. However, since the heuristic feature-extractors embed biases of the human designer regarding visual properties of the target findings, unexpected patterns are not well detected severely constraining the performance in real world applications. CNN for segmentation~\cite{chen2016deeplab} or detection~\cite{ren2015faster} would improve the performance, however, manual annotation of lesions is labor-intensive especially when they are spread in the images, thus, renders the process of data collection highly expensive.

In this paper, we demonstrate an inexpensive and efficient approach to collecting regional annotation of findings and propose a CNN architecture that classifies the existence of a target finding and localizes the lesions. We show that training with the guidance of regional cues not only helps localize the lesions of findings more precisely but also improves classification performance in some cases. This is possible because the regional guidance encourages the network to learn right patterns of findings instead of biases in the images.

\section{Proposed Methods}
\subsection{Data Collection}
\begin{figure}[!th]
  \centering
\includegraphics[scale=0.29]{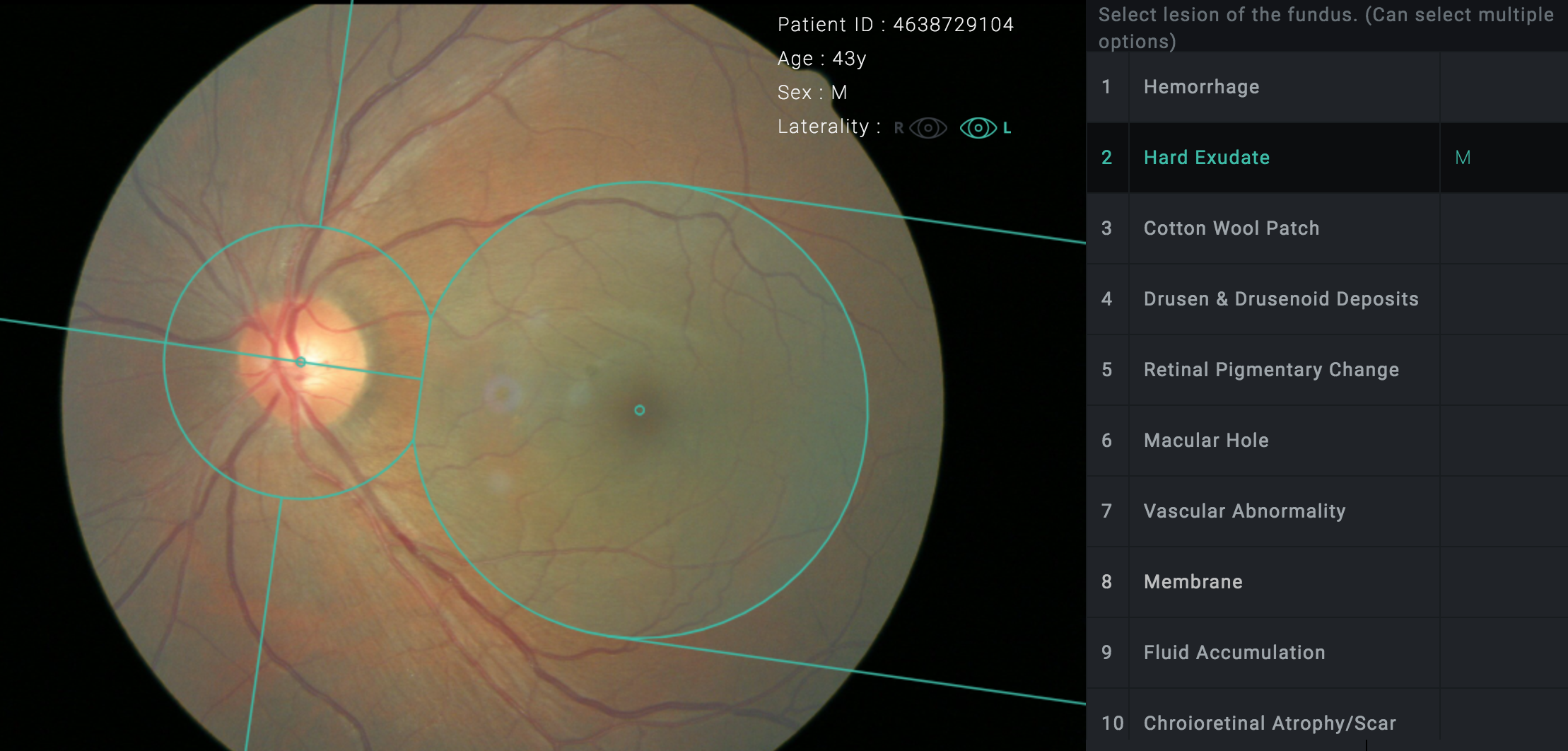}
\caption{Data collection system developed to retrieve regional annotations of findings in fundoscopic images.}
\label{fig:data_collection}
\end{figure}
We collected regional annotations of findings for macular-centered images obtained at health screening centers and outpatient clinics in Seoul National University Bundang Hospital using a data collection system (Fig.~\ref{fig:data_collection}). Annotators chose a type of finding on the right panel and selected the corresponding regions on the left panel. When the eye was normal, no finding was annotated to the image.

We divided an image into 8 regions in a way that each region reflects the anatomical structure of the eyes and the regional characteristics of findings. When the distance between the optic disc and fovea is $D$, circles are drawn at the centers of the optic disc and fovea with the radius of $\frac{2}{5}D$ and $\frac{2}{3}D$ and the intersections of the two circles are connected with a line segment. Then, a half-line passing through the optic disc and fovea ($L$) cuts the circle of the optic disc in half and two half-lines parallel to $L$ and tangent to the circle of fovea are drawn in a direction away from the optic disc. Finally, a line perpendicular to $L$ is drawn to pass through the center of the optic disc.

We also separated annotations into training and test sets based on the expertise of the annotators. Training set was annotated by 27 board-certified ophthalmologists and the test set was annotated by 16 certified retina specialists and 9 certified glaucoma specialist. Each fundoscopic image was annotated by 3 ophthalmologists in total. Training and test set amount to 66,473 and 15,451 images respectively. This study was approved by the institutional review board at Seoul National Bundang Hospital (IRB No. B-1508-312-107) and conducted in accordance with the tenets of the Declaration of Helsinki. 

\subsection{Network Architecture}
\begin{figure}[!th]
  \centering
\includegraphics[scale=0.33]{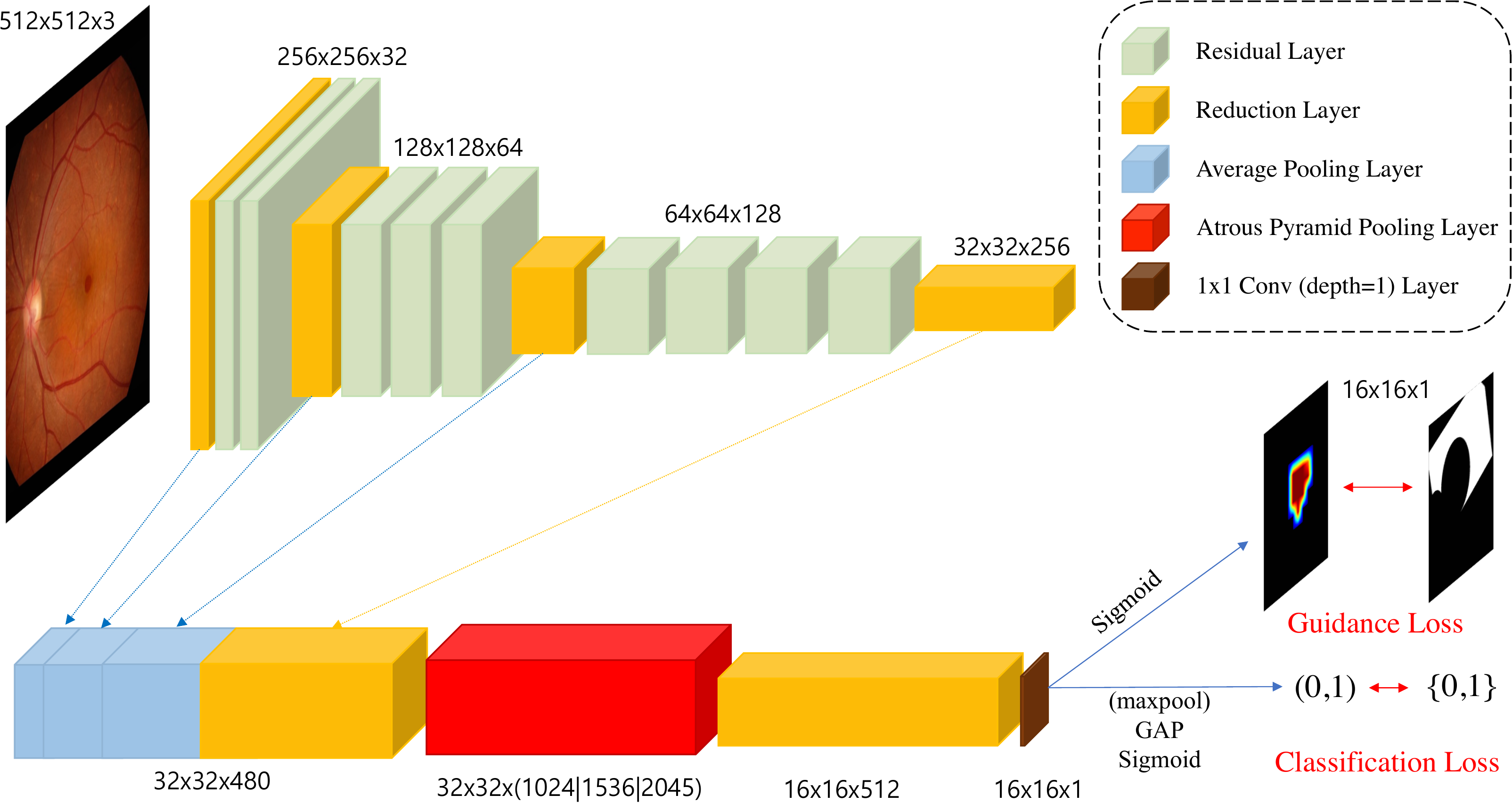}
\caption{Proposed network architecture for localization and classification of a specific finding.}
\label{fig:network}
\end{figure}

As shown in Fig.~\ref{fig:network}, our network architecture consists of residual layer (feature maps after residual unit~\cite{he2016deep}), reduction layer (feature maps after $3\times 3$ conv with stride 2, batch-norm, ReLU), average pooling layer, atrous pyramid pooling layer~\cite{chen2016deeplab} and $1\times 1$ conv (depth=1) layer. 

The depth of layers doubles when the height and width halve. First four reduction layers with different sizes are concatenated after average pooling to exploit both low level and high level features. The concatenated feature maps are atrous-pyramid-pooled with the dilation rates of 1,2,4,8 (findings with large size), or the rates of 1,2,4 (findings with medium size) or the rates of 1,2 (findings with small size). We employed atrous pyramid pooling to aggregate features with various size of receptive fields~\cite{chen2016deeplab}.

Note that the $1\times 1$ conv layer is a linear combination of the previous feature maps as in class activation map(CAM)~\cite{zhou2016learning}. With a sigmoid function, values in the layer are normalized to (0,1), thus, the consequent layer can be considered as normalized activation map. Our activation map differs from CAM in that additional loss function guides the activation to appear only in desirable areas. Also, the $1\times 1$ conv layer is globally average-pooled and normalized with sigmoid function to yield the prediction regarding the existence. Therefore, the activation map is directly related to the prediction in our network architecture and does not require external operations for visualization. In case of findings with small lesions (hemorrhage, hard exudate, cotton wool patch, drusen and retinal pigmentary change), max-pooling layer is inserted before the GAP layer to compensate for the low GAP values to compensate for low GAP values.

\subsection{Objective Function}
For a fundoscopic image $(I\in R^{W_I\times H_I})$, the existence of a target finding in the image $I$ is encoded to $y_{true}\in\{0,1\}$ and the probability of the existence $y_{pred}\in(0,1)$ is the output from the network. When $k$ images are given as a mini-batch, binary cross entropy for classification loss in Fig.~\ref{fig:network} is given by
\begin{equation}
L_{class}({\bf y}_{true},{\bf y}_{pred})=\frac{1}{k}\sum^{k}_{i=1}\big[-y^i_{true} \log y^i_{pred} - (1-y^i_{true})\log (1-y^i_{pred})\big]
\label{loss:classification}
\end{equation}
where ${\bf y}_{true}=\{y^1_{true},...,y^k_{true}\}$ and ${\bf y}_{pred}=\{y^1_{pred}, ...,y^k_{pred} \}$.

When the last feature maps are size of $W_F\times H_F$, a region mask for a target finding $(M\in \{0,1\}^{W_F\times H_F})$ is given as label and the activation map $(A\in(0,1)^{W_F\times H_F})$ is generated from the network. With a mini-batch of size $k$, guidance loss in Fig.~\ref{fig:network} is given by 
\begin{equation}
L_{guide}({\bf A},{\bf M})=\frac{1}{kW_FH_F}\sum^{k}_{i=1} \sum^{W_F H_F}_{l=1} (1-m^i_l)\log (\max(a^i_l, \epsilon))
\label{loss:guidance}
\end{equation}
where ${\bf A}=\{A^1,...,A^k\}$ and ${\bf M}=\{M^1,...,M^k\}$ and $m^i_l$ and $a^i_l$ are values at $l$th pixel in $M^i$ and $A^i$ for $l=1,...,W_FH_F$. Note that $\epsilon>0$ is added inside the logarithm for numerical stability when $a^i_l\approx 0$. In a nutshell, the guidance loss suppresses any activation $(a^i_l)$ in regions where the value of the mask is 0 $(m^i_l=0)$ and has no effect for activation inside the mask $(m^i_l=1)$.

Then, total loss is given by combining the classification loss and the guidance loss,
\begin{equation}
L_{total}= L_{class}({\bf y}_{true}, {\bf y}_{pred}) + \lambda L_{guide}({\bf A},{\bf M})
\label{loss}
\end{equation}
where $\lambda$ balances two objective functions. 

\section{Experiments}
\subsection{Experimental Setup}
We selectively show results of clinically important findings - findings associated with DR and DME~\cite{wilkinson2003proposed} (hemorrhage, hard exudate, drusen, cotton wool patch (CWP)), macular hole, membrane and retinal nerve fiber layer defect (RNFL defect). The training set is split into derivation set ($90\%$) and validation set ($10\%$). Model was optimized with the derivation set until the validation loss stagnates and exacerbates. The model with the lowest validation loss is tested on the test set which we regard as gold standards. We defined that a target finding is absent when no ophthalmologists annotated and present when more than 2 out of 3 ophthalmologists annotated. The union of annotated regions is provided as regional cues during training. 

We aim to measure the effectiveness of the guidance loss by experimenting with our CNN architecture (Fig.~\ref{fig:network}) with/without the regional guidance and comparing the results in terms of Area Under Receiver Operating Characteristic curve (AUROC), specificity, sensitivity and activations in the regional cues (AIR). AIR is defined as the summation of activations inside the regional cues divided by the summation of all activations. AIR is measured for both true positive and false negative in classification where the regional cues are available. We used the same network architecture (Fig.~\ref{fig:network}) to implement the networks with or without the regional guidance by changing $\lambda$ in Eq.\ref{loss} ($\lambda=0$ without the guidance).

Original color fundoscopic images are cropped to remove black background and resized to $512\times 512$ for the network input. The resized images are randomly augmented by affine transformation (flip, scaling, rotation, translation, shear) and random re-scaling of the intensity. An image is normalized to $[0,1]$ by dividing by 255. Weights and biases are initialized with Xavier initialization~\cite{glorot2010understanding}. As an optimizer, we used SGD with {\it nesterov} momentum $0.9$ and decaying learning rate. Batch size is set to 32 following the recommendation that small batch size leads to better generalization~\cite{keskar2016large}. We set $\epsilon =10^{-3}$ in Eq.\ref{loss:guidance} to obtain numerical stability and $\lambda=1$ in Eq.\ref{loss} to treat classification loss and guidance loss equally.

\subsection{Experimental Results}
Comparison of the performance between inception-v3~\cite{gulshan2016development} and our two models (with/without guidance loss) is summarized in Table~\ref{tab:result}. We can observe positive effects of the guidance loss to AIR for TP and FN throughout all findings. This is desirable, since it means that the network attends inside the regional cues for classification, thus the network is less likely to learn biases of datasets. Also, difference in AIR is larger in the cases of TP between the two models than those of FN. This is reasonable since FN consists of hard cases for the networks to classify, while TP is relatively easy to be classified with high confidence.

\begin{table}[!th]
\centering
\caption{Comparison of inception-v3 and our two models (with/without the guidance) on the test set with respect to AUROC. Activations in the regional cues (AIR) on true positive (TP) and false negative (FN) in classification. Among multiple operating points, specificity and sensitivity that yield the best harmonic mean are chosen. }
\resizebox{\columnwidth}{!}{
\begin{tabular}{C{9em} |C{6em} C{4.2em} C{4.2em}| C{4.2em} C{4.2em} |C{4.2em} C{4.2em}}
  \toprule
  \multirow{2}{*}{Findings}& \multicolumn{3}{c|}{AUROC}& \multicolumn{2}{c|}{AIR (TP)}&\multicolumn{2}{c}{AIR (FN)}\\ & Inception-v3 & With & Without  &With  & Without &With & Without \\
  \midrule
  Macular Hole & 0.9592&{\bf 0.9870}& 0.9676  & {\bf 0.9999}& 0.3156& {\bf 0.9999}& 0.2611\\
  Hard Exudate & 0.9889 &{\bf 0.9938}&  0.9910& {\bf 0.8089}& 0.5999&{\bf 0.5750}& 0.4614\\
  Hemorrhage & 0.9760 & 0.9862 & {\bf  0.9895}& {\bf 0.8890}& 0.6388&{\bf 0.4899}& 0.3857\\
  Membrane & 0.9654&{\bf 0.9831}& 0.9795 &{\bf 0.9699}& 0.3696&{\bf 0.8446}& 0.2499\\
  Drusen& 0.9746&{\bf 0.9811}& 0.9786& {\bf 0.8292}& 0.5611& {\bf 0.5931}& 0.4012\\
  Cotton Wool Patch& 0.9633&{\bf 0.9792}&  0.9741 & {\bf 0.8058}& 0.5450&{\bf 0.4672}& 0.4538\\
  RNFL Defect& 0.9037& {\bf 0.9263}&  0.8870&{\bf  0.7233}& 0.4024& {\bf 0.4801}& 0.2838\\
  \bottomrule
\end{tabular}
} 
\label{tab:result}
\end{table}

When it comes to AUROC, only macular hole and RNFL defect showed significant improvements. It is interesting to notice that these findings are observed in specific regions. This can be explained by the fact that learning becomes easier as the network is guided to attend to important regions for classification. On the other hand, findings that spread over the extensive areas such as hemorrhage, hard exudate and drusen took less or no advantage of regional cues for classification. We suspect that this happens because these findings would have wide regional cues that the guidance is marginal and the lesions are small that guidance would be more difficult. It is observed that when AUROC is higher, sensitivity is also higher and specificity is lower. However, significant difference is seen only for macular hole and RNFL defect.

\begin{figure}[!th]
\captionsetup[subfigure]{labelformat=empty}
  \centering
\hspace{-2mm}
\subfloat{\includegraphics[width=3cm,height=3cm]{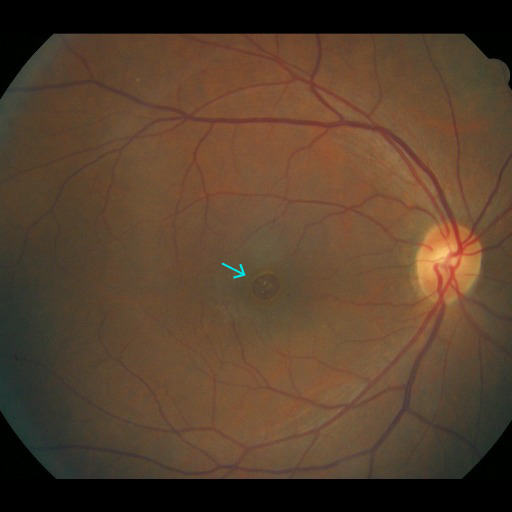}}\hspace{0mm}
\subfloat{\includegraphics[width=3cm,height=3cm]{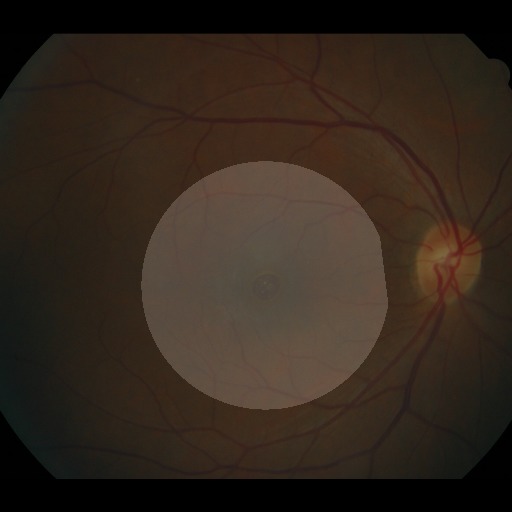}}\hspace{0mm}
\subfloat{\includegraphics[width=3cm,height=3cm]{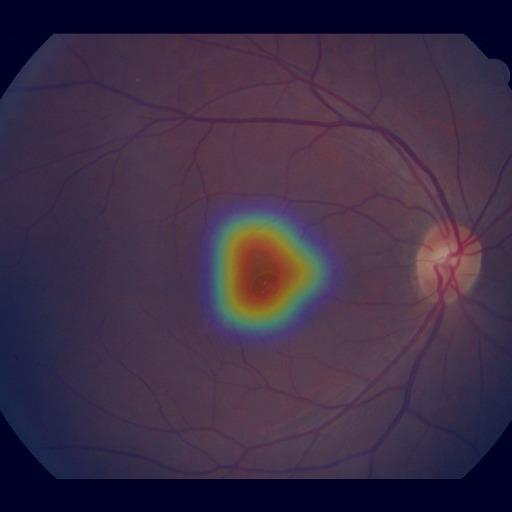}}\hspace{0mm}
\subfloat{\includegraphics[width=3cm,height=3cm]{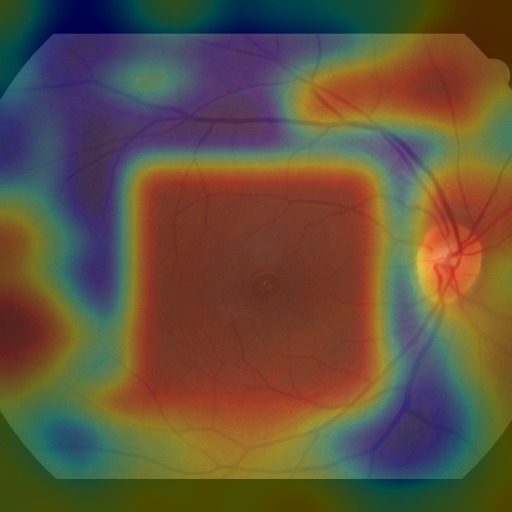}}\hspace{0mm}\\
\hspace{-2mm}
\subfloat{\includegraphics[width=3cm,height=3cm]{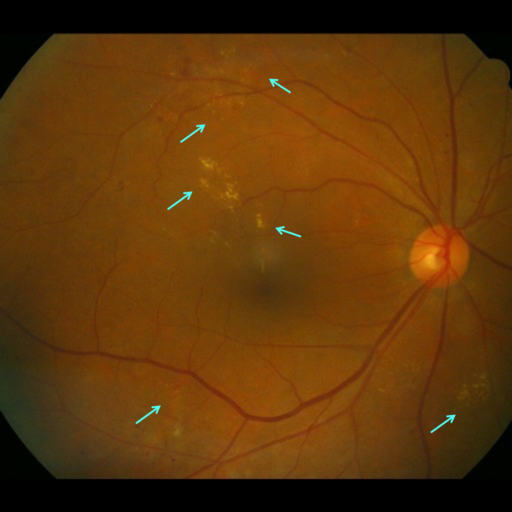}}\hspace{0mm}
\subfloat{\includegraphics[width=3cm,height=3cm]{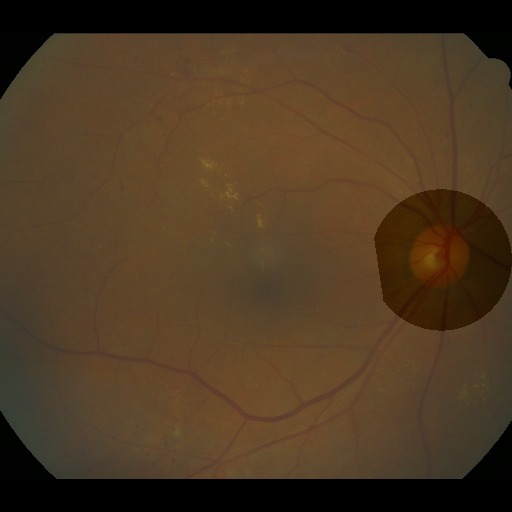}}\hspace{0mm}
\subfloat{\includegraphics[width=3cm,height=3cm]{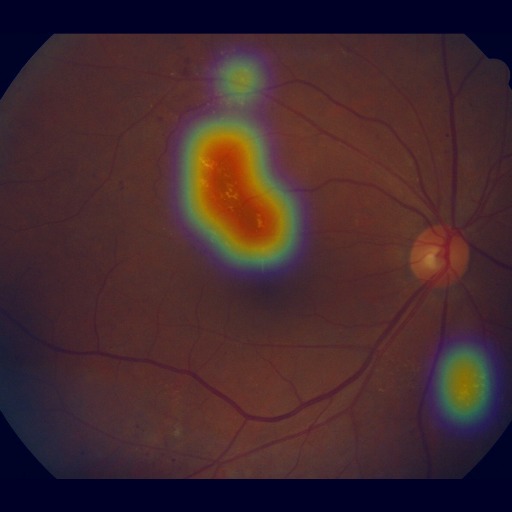}}\hspace{0mm}
\subfloat{\includegraphics[width=3cm,height=3cm]{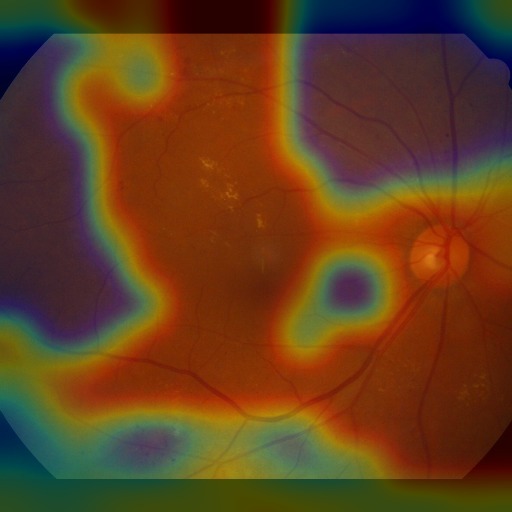}}\hspace{0mm}\\
\hspace{-2mm}
\subfloat{\includegraphics[width=3cm,height=3cm]{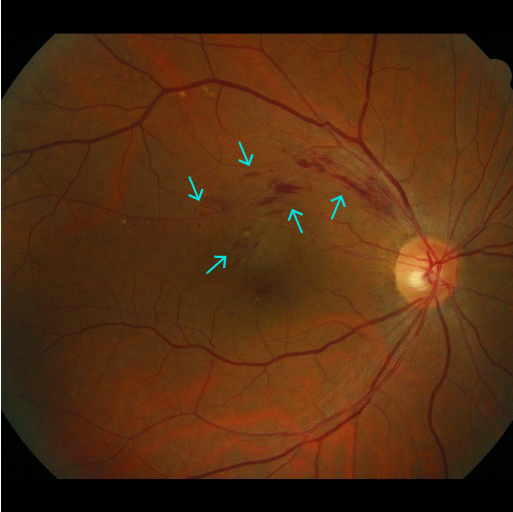}}\hspace{0mm}
\subfloat{\includegraphics[width=3cm,height=3cm]{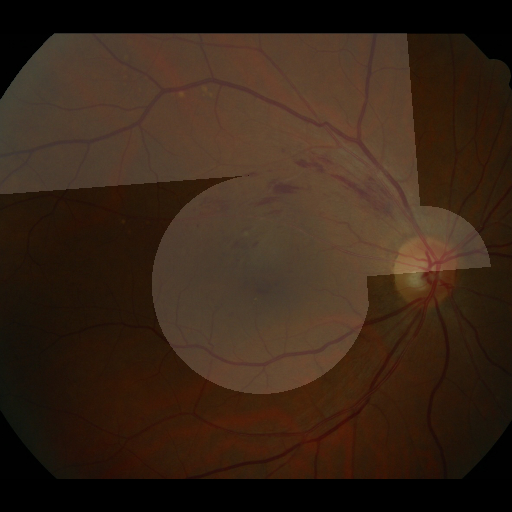}}\hspace{0mm}
\subfloat{\includegraphics[width=3cm,height=3cm]{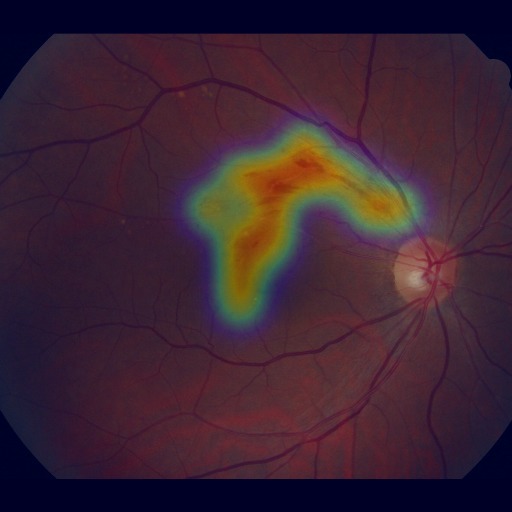}}\hspace{0mm}
\subfloat{\includegraphics[width=3cm,height=3cm]{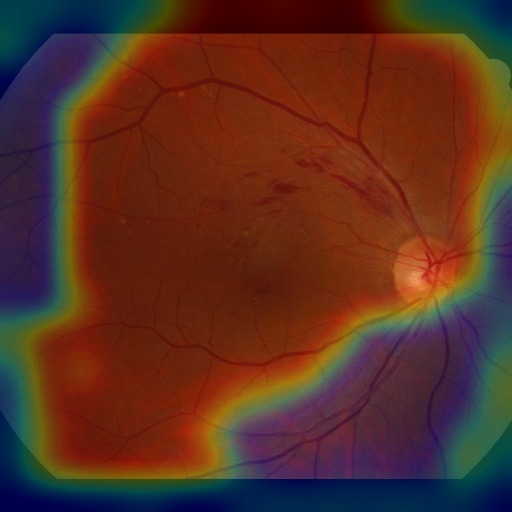}}\hspace{0mm}\\
\hspace{-2mm}
\subfloat{\includegraphics[width=3cm,height=3cm]{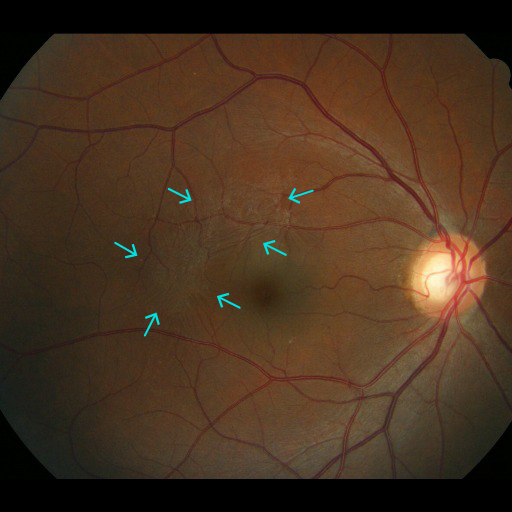}}\hspace{0mm}
\subfloat{\includegraphics[width=3cm,height=3cm]{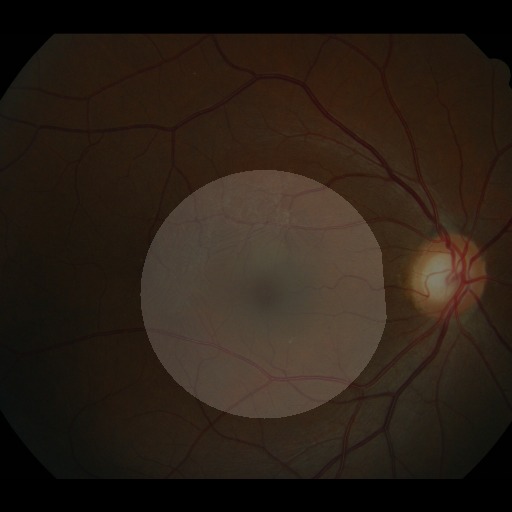}}\hspace{0mm}
\subfloat{\includegraphics[width=3cm,height=3cm]{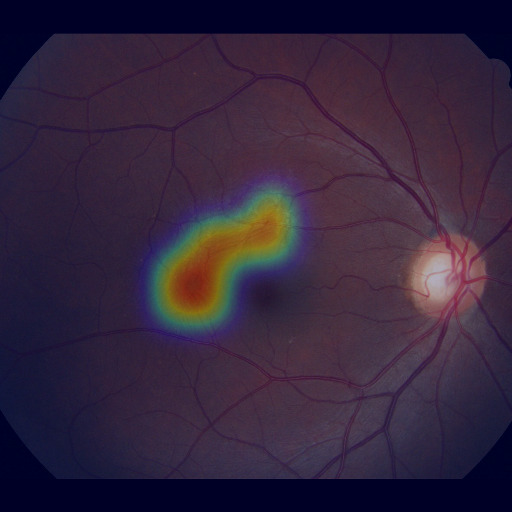}}\hspace{0mm}
\subfloat{\includegraphics[width=3cm,height=3cm]{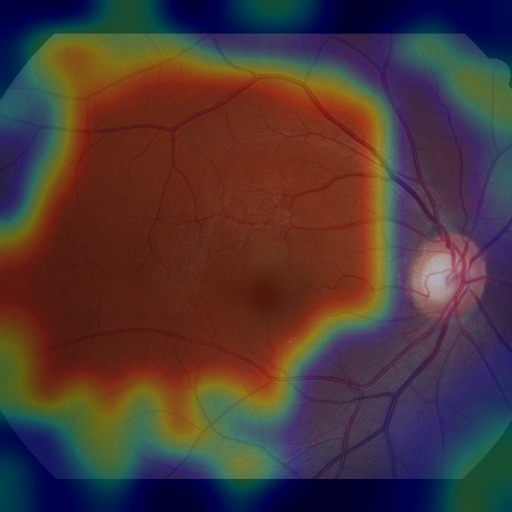}}\hspace{0mm}\\
\hspace{-2mm}
\subfloat{\includegraphics[width=3cm,height=3cm]{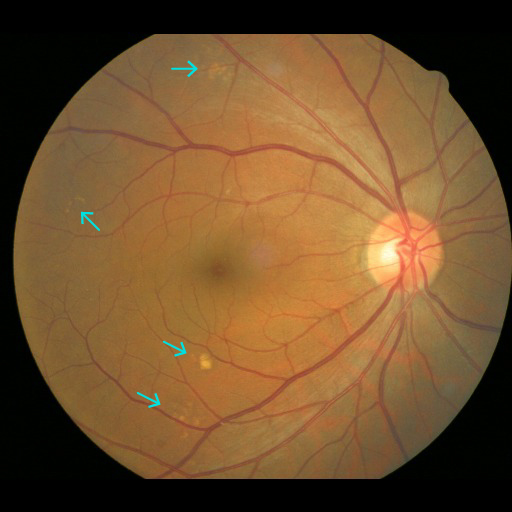}}\hspace{0mm}
\subfloat{\includegraphics[width=3cm,height=3cm]{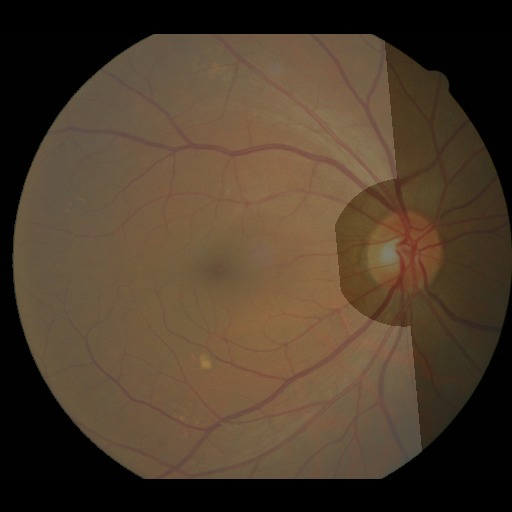}}\hspace{0mm}
\subfloat{\includegraphics[width=3cm,height=3cm]{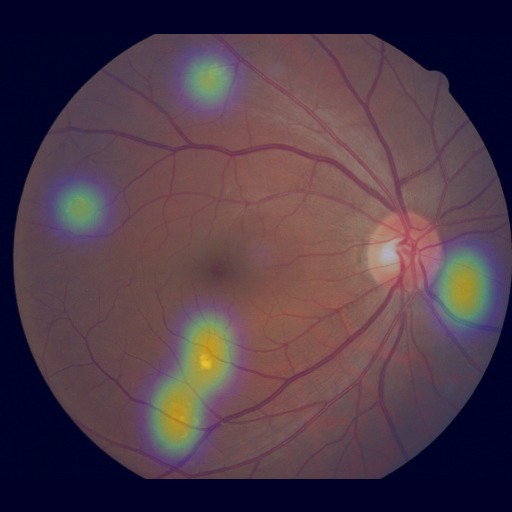}}\hspace{0mm}
\subfloat{\includegraphics[width=3cm,height=3cm]{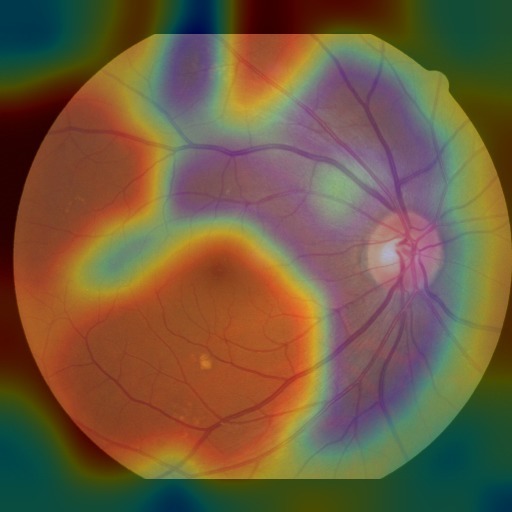}}\hspace{0mm}\\
\hspace{-2mm}
\subfloat{\includegraphics[width=3cm,height=3cm]{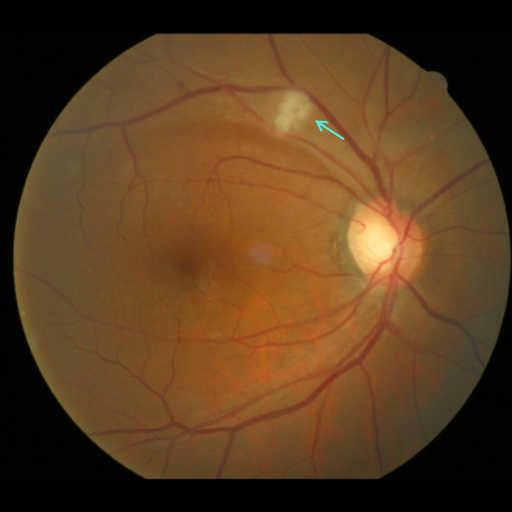}}\hspace{0mm}
\subfloat{\includegraphics[width=3cm,height=3cm]{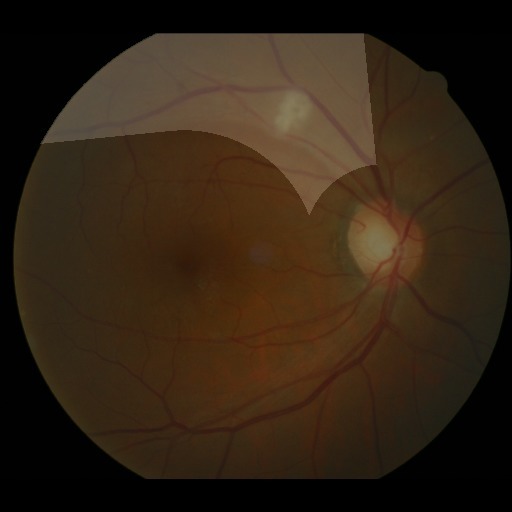}}\hspace{0mm}
\subfloat{\includegraphics[width=3cm,height=3cm]{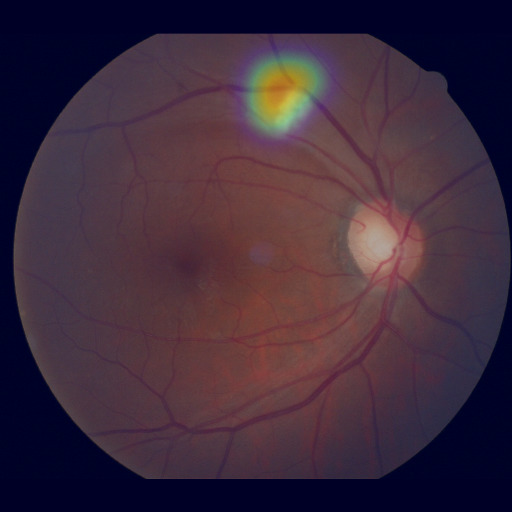}}\hspace{0mm}
\subfloat{\includegraphics[width=3cm,height=3cm]{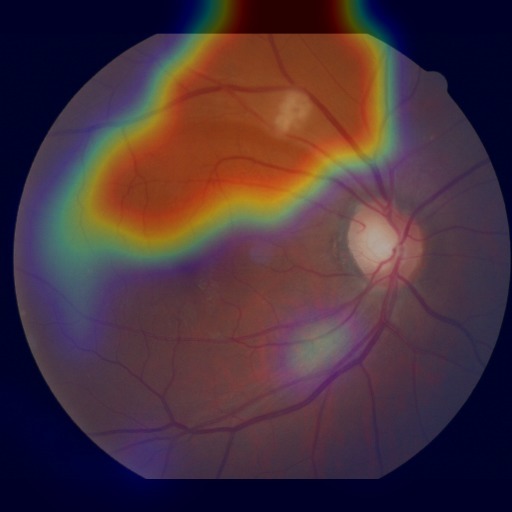}}\hspace{0mm}\\
\hspace{-2mm}
\subfloat{\includegraphics[width=3cm,height=3cm]{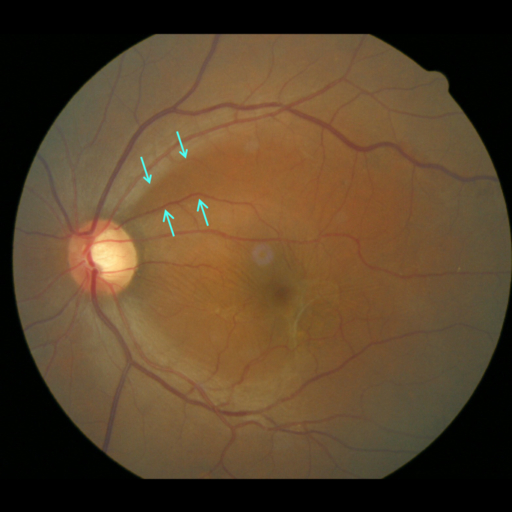}}\hspace{0mm}
\subfloat{\includegraphics[width=3cm,height=3cm]{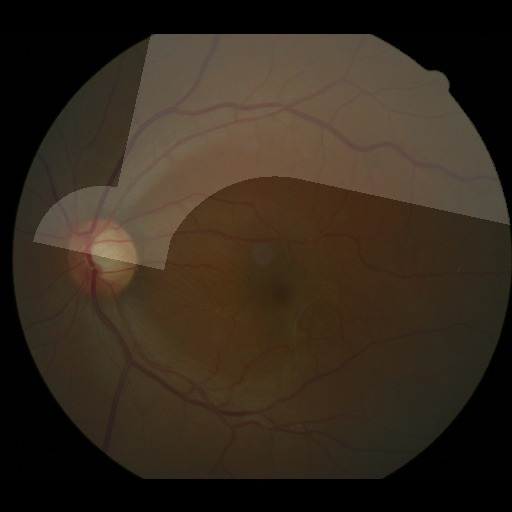}}\hspace{0mm}
\subfloat{\includegraphics[width=3cm,height=3cm]{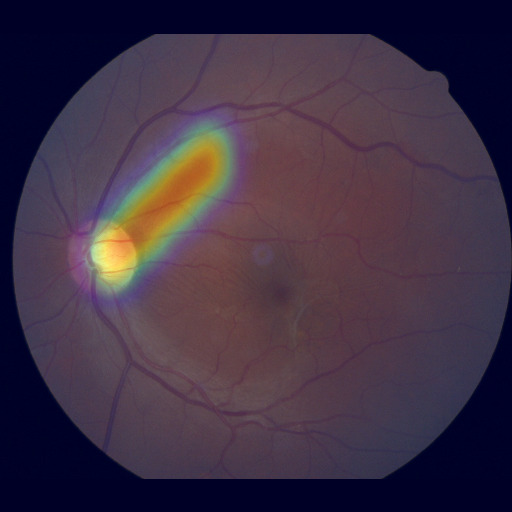}}\hspace{0mm}
\subfloat{\includegraphics[width=3cm,height=3cm]{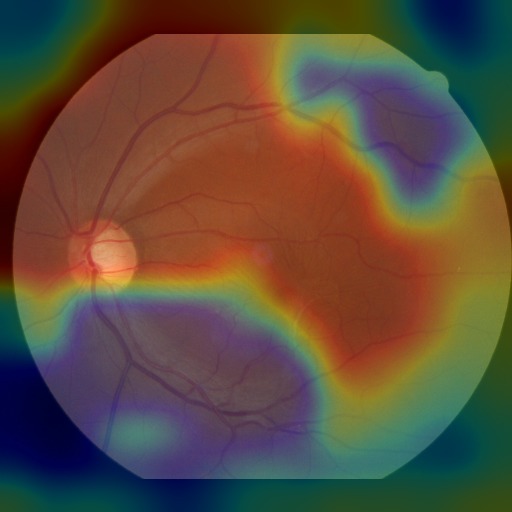}}\hspace{0mm}\\
\caption{({\bf From left to right}) original fundus image, mask, activation map with and without guidance loss. ({\bf From top to bottom}) Macular Hole, Hard Exudate, Hemorrhage, Membrane, Drusen, Cotton Wool Patch, RNFL Defect.}
\label{fig:activation_map}
\end{figure}

In Fig~\ref{fig:activation_map}, we qualitatively compare activation maps of networks with/without the guidance loss. Before superimposed onto the original image, activation maps are upscaled through bilinear interpolation and blurred with $32\times 32$ Gaussian filter for natural visualization and normalized to $[0,1]$. As obvious in the figure, the network generates much more precise activation maps when trained with the regional cues. Though it is not as salient as is segmented pixelwise and includes few false positives in some cases, our activation maps provide meaningful information about the location of the findings which would be beneficial to clinicians. Without the guidance loss, activation maps span far more than the surroundings of lesions and sometimes highlight irrelevant areas.

\section{Conclusion and Discussion}
In this paper, we introduced an approach to exploiting regional information of findings in fundoscopic images for localization and classification. We developed an efficient labeling tool to collect regional annotations of findings and proposed a network architecture that classifies findings with localization of the lesions. When trained with the guidance loss that makes use of the regional cues, our network generates more precise activation maps with better attention to the relevant areas for classification. Also, the proposed regional guide also improves the classification performance of findings that occur only at specific regions. 

\clearpage
\bibliographystyle{splncs03}
\bibliography{paper4}
\end{document}